\documentclass{article}
\usepackage[T1]{fontenc}
\usepackage{iclr2027_conference,times}

\usepackage{amsmath,amsfonts,bm}

\def\eqref#1{equation~\ref{#1}}
\def\1{\bm{1}}

\DeclareMathAlphabet{\mathsfit}{\encodingdefault}{\sfdefault}{m}{sl}
\SetMathAlphabet{\mathsfit}{bold}{\encodingdefault}{\sfdefault}{bx}{n}

\usepackage{graphicx}
\usepackage{booktabs}
\usepackage{wrapfig}
\usepackage{needspace}
\usepackage{hyperref}
\usepackage{url}


\title{Multitask Adaptation of World Model with Capabilities Retention}

\author{%
Zeyu Shen\textsuperscript{1}\thanks{Corresponding author: \href{mailto:zshen39@jh.edu}{\texttt{zshen39@jh.edu}}}%
\quad Zheyuan Zhang\textsuperscript{2}%
\quad Yifan Yin\textsuperscript{2}%
\quad Jennie Si\textsuperscript{3}\\[0.5em]
{\normalfont\small \textsuperscript{1}Data Science and AI Institute, Johns Hopkins University}\\
{\normalfont\small \textsuperscript{2}Department of Computer Science, Johns Hopkins University}\\
{\normalfont\small \textsuperscript{3}School of Electrical, Computer and Energy Engineering, Arizona State University}%
}
\iclrfinalcopy
\hypersetup{
  pdftitle={Multitask Adaptation of World Model with Capabilities Retention},
  pdfauthor={Zeyu Shen, Zheyuan Zhang, Yifan Yin, Jennie Si}
}

\begin{document}

\maketitle

\begin{abstract}
Recent advances in world modeling have enabled generalist agents to learn
diverse tasks across embodiments within a single model, suggesting a path
toward foundation models for continuous control. However, joint training on
all desired tasks is not always feasible, as limited resources
constrain access to collected data and simulation environments at a given
time. An important goal is therefore to adapt trained multitask agents to
newly accessible tasks while retaining their existing capabilities. We propose
MACR, short for \textbf{M}ultitask world model \textbf{A}daptation with \textbf{C}apability \textbf{R}etention, a framework for adapting a pretrained multitask world model to multiple continuous control tasks simultaneously.
Given a short probing trajectory, our framework retrieves a task-relevant
dynamics representation from the trained world model to initialize
adaptation, enabling the agent to learn multiple new tasks efficiently. We
adapt world-model weights by jointly optimizing for new-task learning and
capability retention. A factorized parameterization expresses the adapted
weights as existing or newly learned combinations of fixed pretrained bases,
while comprehensive component-wise distillation uses stored exemplars to
regularize the model's functional behavior when conditioned on previously
learned tasks. Experiments across diverse continuous control tasks
show that our approach learns new tasks while retaining performance on
previously learned tasks and outperforms a range of existing baselines.
These results support MACR as a promising approach to
broadening the capabilities of a multitask world model.
\end{abstract}

\section{Introduction}
\label{sec:intro}

Building general-purpose models for continuous control has become an
increasingly important goal in robot learning. Substantial progress has been
made along two complementary directions: imitation learning and reinforcement
learning (RL). Imitation-learning approaches train large control policies on
broad collections of expert trajectories spanning multiple tasks and
embodiments~\citep{reed2022gato,brohan2022rt1,brohan2023rt2,octo2024,kim2024openvla}.
Their performance depends on the quality and coverage of the collected
demonstrations, while obtaining diverse, high-quality demonstrations can be
costly~\citep{mandlekar2022robomimic,khazatsky2024droid}.
In parallel, RL learns control policies through interaction and reward
feedback. Model-free multitask methods, including extensions of soft
actor-critic (SAC) and proximal policy optimization (PPO), learn shared
policies across tasks~\citep{yu2019metaworld,joshi2025mtbench}, but can require
substantial environment interaction and suffer from interference between task
objectives~\citep{yu2020pcgrad}. Model-based methods can improve data efficiency
by planning with learned dynamics, but face challenges from model inaccuracies
and the computational cost of long-horizon planning~\citep{hansen2022tdmpc}.
Recent world-model methods integrate learned latent dynamics with policy and
value learning, combining model-based prediction with model-free components.
The Dreamer family optimizes behavior through imagined rollouts in a learned
latent model, whereas the TD-MPC family integrates latent world models with
temporal-difference learning and model-predictive
control~\citep{hafner2020dreamer,hafner2025dreamerv3,hansen2022tdmpc,hansen2024tdmpc2}.
Recent multitask extensions further demonstrate broad multitask learning and
favorable scaling with model and data size~\citep{hansen2024tdmpc2,hansen2025newt}.
Together, these efforts point toward foundation models for continuous control:
a single trained agent can support a broad range of tasks.

% Declare the float after introductory text so the compact figure stays on page 2.
\begin{figure}[t]
  \centering
  \includegraphics[width=\linewidth]{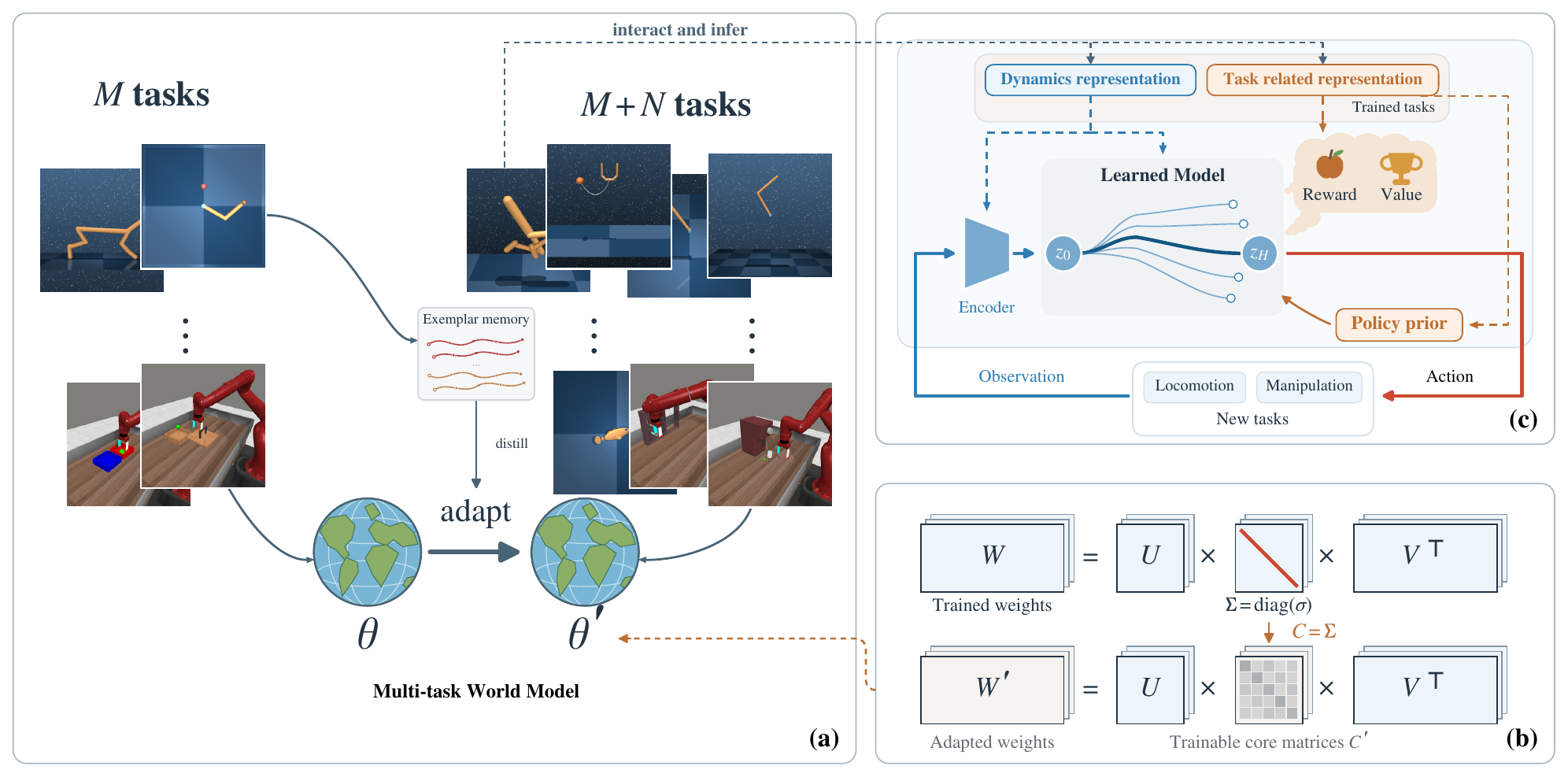}
  \caption{\textbf{Overview of our framework.}
  (a) A trained multitask world model learns new tasks while retaining prior
  capabilities through exemplar-based distillation.
  (b) Adaptation learns core weight matrices within fixed SVD bases.
  (c) Separate retrieval of dynamics and task-related representations transfers
  knowledge for prediction and control.}
  \label{fig:method_overview}
\end{figure}

Despite this progress, existing generalist agents are predominantly trained
with a fixed number of tasks defined in advance. Once training is complete,
extending the model to a new task is typically treated as a separate
transfer-learning problem. Fine-tuning is a common way to adapt a trained
model to a held-out task, but it may damage performance on previously learned
tasks~\citep{mccloskey1989catastrophic,french1999catastrophic}.
Moreover, prior work on learning new tasks while preserving existing
capabilities has largely considered tasks that arrive one at a
time~\citep{khetarpal2022continualrl,wolczyk2021continualworld}, leaving
simultaneous adaptation to multiple new tasks with retention of previously
learned capabilities comparatively underexplored.

Advances in model scaling have enabled larger control models to learn diverse
tasks jointly~\citep{hansen2024tdmpc2}. In parallel, research on replay,
regularization, architecture design, optimization, and representation learning
has developed complementary mechanisms for retaining previously learned
capabilities~\citep{rolnick2019experience,kirkpatrick2017ewc,rusu2016progressive,lopezpaz2017gem,cha2021co2l,wang2024comprehensive}.
Building on these advances, we propose MACR, a framework for multitask world
model adaptation with capability retention. Our main contributions are as
follows:
(1) By briefly exploring each new task, MACR retrieves dynamics-related
and task-related representations from a trained multitask world model,
providing an effective initialization for adaptation. It then jointly learns
multiple new tasks by optimizing core weight matrices within fixed SVD bases,
while exemplar-based distillation helps retain prior knowledge and skills.
(2) We conduct an exploratory empirical study of MACR for simultaneous
adaptation to multiple continuous control tasks across
DMControl~\citep{tassa2018dmcontrol} and Meta-World~\citep{yu2019metaworld}.
These results suggest that multitask adaptation with capability retention
is a promising way to extend the capabilities of generalist agents, which
are typically trained across embodiments and control tasks.

\section{Related Work}
\label{sec:related-work}

\paragraph{Multitask RL.}
Multitask reinforcement learning aims to share representations and parameters
across tasks within a single agent. Model-free approaches improve task sharing
through gradient conflict mitigation, modular policies, and context-conditioned
representations~\citep{yu2020pcgrad,yang2020softmodularization,sodhani2021care}.
Model-based methods learn environment dynamics and use predicted trajectories
for planning~\citep{chua2018pets}. Within this family, multitask world-model
agents such as TD-MPC2 and Newt combine model-based prediction and planning
with model-free policy and value learning, supporting shared control across
diverse tasks and embodiments~\citep{hansen2022tdmpc,hansen2024tdmpc2,hansen2025newt}.

\paragraph{Scaling RL for Control Foundation Models.}
Large-scale imitation learning has enabled generalist control policies such as
RT-1~\citep{brohan2022rt1} and Octo~\citep{octo2024}. Collecting expert
demonstrations for new tasks remains costly, and the behaviors learned through
imitation depend on the quality and coverage of those
demonstrations~\citep{khazatsky2024droid,hansen2025newt}. Reinforcement learning
offers a complementary route by using reward feedback to improve behavior, as
demonstrated in game playing~\citep{silver2018alphazero} and language model
reasoning~\citep{guo2025deepseekr1}. In continuous control, TD-MPC2 trained a
single world model on 80 tasks across DMControl and Meta-World using offline
experience of varying quality, with performance improving as model capacity
increased~\citep{hansen2024tdmpc2}. Scaling online RL across diverse tasks
also requires stable optimization~\citep{hansen2025newt} and sufficient
environment interaction throughput~\citep{weng2022envpool}. These advances
suggest that multitask RL can provide a basis for control foundation models.
We show that a pretrained world model can adapt jointly to multiple new tasks
with limited online interaction while retaining previously acquired capabilities.

\paragraph{Adaptation and Knowledge Retention.}
A straightforward approach to adaptation is to fine-tune a pretrained model
on experience collected from new tasks. Meta-learning methods such as MAML
learn initializations that facilitate rapid
fine-tuning~\citep{finn2017maml,kaushik2020famle}, while context-based approaches
infer task variables or dynamics parameters from interaction histories to
condition control~\citep{rakelly2019pearl,xian2021hyperdynamics,zhou2025certain}.
These adaptation formulations primarily target performance on new tasks,
with retention of previously acquired capabilities generally outside their
objectives. Continual learning aims to mitigate catastrophic forgetting when
learning new tasks. Existing approaches include
replay~\citep{shin2017continual,rolnick2019experience},
regularization~\citep{kirkpatrick2017ewc,zenke2017continual,aljundi2018memory},
architecture design~\citep{rusu2016progressive,mallya2018packnet}, and
optimization~\citep{lopezpaz2017gem}. Much of this literature assumes that tasks
arrive sequentially, one at a time~\citep{wang2024comprehensive}. Sequential
adaptation may limit efficiency when multiple new tasks become available
together and can be learned through joint optimization with shared experience.
MACR addresses joint adaptation of a pretrained multitask world model to
multiple new tasks while retaining its previously acquired capabilities.

\section{Multitask Adaptation with Capability Retention}
\label{sec:online-multitask-incremental-learning}

\subsection{Problem Formulation}
\label{sec:background}
\label{sec:problem-formulation}

An agent learns a control policy through interaction with an environment.
We model each task as a discounted Markov decision process
\(\mathcal{M}_i=(\mathcal{S}_i,\mathcal{A}_i,p_i,r_i,\gamma_i)\),
where \(\mathcal{S}_i\) and \(\mathcal{A}_i\) are the state and action spaces,
\(p_i(s'\mid s,a)\) is the transition distribution, \(r_i(s,a)\) is the reward
function, and \(\gamma_i\) is the discount factor. The learning
objective is to find a policy \(\mu(a\mid s,i)\) that maximizes the expected
discounted return
\(J_i(\mu)=\mathbb{E}_{\mu}[\sum_{t=0}^{\infty}\gamma_i^t r_i(s_t,a_t)]\).

We study multitask adaptation with capability retention. Consider a world
model \(f(\theta,E)\) initially trained on
\(\mathcal{T}_{\mathrm{pre}}=\{\mathcal{M}_1,\ldots,\mathcal{M}_N\}\) and a
set of newly available tasks
\(\mathcal{T}_{\mathrm{new}}=\{\mathcal{M}_{N+1},\ldots,\mathcal{M}_{N+n}\}\).
Without assuming access to pre-collected training datasets or expert
demonstrations, we begin adaptation to each new task
\(\mathcal{M}_i\in\mathcal{T}_{\mathrm{new}}\) with only a short interaction
context \(\mathcal{C}_i\) collected through random exploration, with
transitions and their observed rewards recorded as \((s,a,r,s')\).
We seek to use this limited context to
identify reusable knowledge in the trained model, providing an effective
initialization that accelerates subsequent adaptation. Our objective is to
learn the new tasks while preserving performance on previously learned
tasks, yielding a single agent that performs well across
\(\mathcal{T}=\mathcal{T}_{\mathrm{pre}}\cup\mathcal{T}_{\mathrm{new}}
=\{\mathcal{M}_1,\ldots,\mathcal{M}_{N+n}\}\).

\Needspace{5\baselineskip}
\subsection{Embedding Retrieval Through Multi-task World Models}
\label{sec:method-retrieval}

\label{sec:background-task-conditioned-world-models}
\begingroup
\setlength{\intextsep}{5pt}
\setlength{\columnsep}{11pt}
\setlength{\abovecaptionskip}{4pt}
\begin{wrapfigure}{r}{0.60\linewidth}
  \centering
  \includegraphics[width=\linewidth]{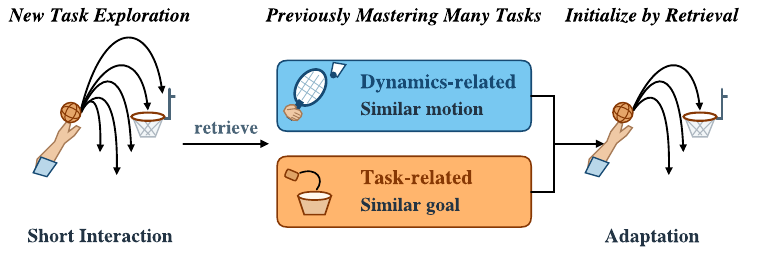}
  \caption{Illustration of Embedding Retrieval}
  \label{fig:split-retrieval-illustration}
\end{wrapfigure}
Following TD-MPC2~\citep{hansen2024tdmpc2}, we consider a world model
comprising an encoder \(h_\theta\), a latent dynamics model \(d_\theta\),
a reward predictor \(R_\theta\), an actor \(\pi_\theta\), and an
action-value function \(Q_\theta\). We group these into dynamics-related
components \((h_\theta,d_\theta)\) and task-related components
\((R_\theta,\pi_\theta,Q_\theta)\). Generally, the agent encodes states into
a latent space and plans using predicted rewards and values along latent
dynamics rollouts, while the actor provides a policy prior to guide action
sampling during planning.
In the multitask setting considered here,
network parameters \(\theta\) are shared across tasks, and each component is
conditioned on a task embedding \(e_j\). The embedding table
\(E=[e_1,\ldots,e_N]^\top\), together with the shared parameters, defines the
task-conditioned model \(f(\theta,E)\).
\par
\endgroup

In a trained multitask model, each task embedding can be regarded as a learned
representation that, together with the shared model weights, determines
task-specific dynamics predictions, policy behavior, and reward and value
estimates. We seek to reuse these representations to initialize a new task
from a short interaction context obtained from trajectories sampled during
random exploration. To adapt the world model to new tasks, we retrieve
embeddings separately for the dynamics-related components
\((h_\theta,d_\theta)\) and the task-related components
\((R_\theta,\pi_\theta,Q_\theta)\), using these representations to reuse
capabilities learned on previous tasks.
Figure~\ref{fig:split-retrieval-illustration} illustrates how the same
interaction context can lead to different embeddings for the two groups.

We score candidate embeddings using the latent consistency, reward
prediction, and temporal-difference losses from TD-MPC2's learning
objective~\citep{hansen2024tdmpc2}, denoted by
\(\mathcal{L}_{\mathrm{dyn}}\), \(\mathcal{L}_R\), and \(\mathcal{L}_Q\),
respectively. Let \(\mathcal{E}=\{e_1,\ldots,e_N\}\) be the set of embeddings
for previously learned tasks. We first use \(\mathcal{C}_i\) to retrieve a
previously learned task embedding for the dynamics-related components and
denote it by \(e^{\mathrm{dyn}}\). Holding \(e^{\mathrm{dyn}}\) fixed, we then
retrieve \(e^{\mathrm{ctrl}}\) for the task-related components:
\[
\begin{aligned}
e^{\mathrm{dyn}}
&\leftarrow
\underset{e_j\in\mathcal{E}}{\arg\min}\;
\mathcal{L}_{\mathrm{dyn}}(e_j;\mathcal{C}_i),\\
e^{\mathrm{ctrl}}
&\leftarrow
\underset{e_j\in\mathcal{E}}{\arg\min}\;
\Big[
\beta_R\mathcal{L}_R(e_j;\mathcal{C}_i)
+
\beta_Q\mathcal{L}_Q(e_j;\mathcal{C}_i)
\Big].
\end{aligned}
\]
Here, \(\beta_R\) and \(\beta_Q\) are the corresponding loss weights in
TD-MPC2's learning objective.

\subsection{Component-Wise Distillation}
\label{sec:method-distillation}

Adapting the shared model to new tasks can alter the representations,
predictions, and policy behavior learned for previous tasks. For a trajectory
\(\tau=(s_t,a_t,r_t,s_{t+1})_{t=0}^{T-1}\) of length \(T\), its undiscounted
episodic return is \(G(\tau)=\sum_{t=0}^{T-1}r_t\). For each original task
\(\mathcal{M}_i\), we retain the \(K\) previously collected trajectories
with the highest returns, denoted by
\(\tau_i^{(1)},\ldots,\tau_i^{(K)}\), in the exemplar memory
\[
\mathcal{B}_{\mathrm{pre}}
=\{(i,\tau_i^{(k)}):i=1,\ldots,N,\;k=1,\ldots,K\}.
\]
The task identity \(i\) is retained for conditioning the model.
During adaptation, the agent does not interact with the environments of
previously learned tasks.

The retained trajectories are paired with the corresponding outputs of the
original model to form exemplars for capability retention. For each exemplar,
we align the adapted encoder's latent state output \(z=h_\theta(s)\) with
the original encoder's output \(\bar z=\bar h(s)\), and the next latent
state prediction \(d_\theta(z,a)\) with \(\bar d(\bar z,a)\).
For the reward predictor \(R_\theta\) and value function \(Q_\theta\),
we match their output distributions to those of the original components
\(\bar R\) and \(\bar Q\).
The original policy distribution \(\bar\pi(\cdot\mid\bar z)\) supplies
policy supervision. The component losses for each exemplar transition are:
\[
\begin{aligned}
\mathcal{L}_h^{\mathrm{dist}}
&=\operatorname{MSE}(z,\bar z),\\
\mathcal{L}_d^{\mathrm{dist}}
&=\operatorname{MSE}\!\left(
d_\theta(z,a),\bar d(\bar z,a)\right),\\
\mathcal{L}_R^{\mathrm{dist}}
&=\operatorname{CE}\!\left(
\bar R(\bar z,a),R_\theta(z,a)
\right),\\
\mathcal{L}_Q^{\mathrm{dist}}
&=\operatorname{CE}\!\left(
\bar Q(\bar z,a),Q_\theta(z,a)
\right),\\
\mathcal{L}_\pi^{\mathrm{dist}}
&=D_{\mathrm{KL}}\!\left(
\bar\pi(\cdot\mid\bar z)\,\|\,
\pi_\theta(\cdot\mid\bar z)
\right).
\end{aligned}
\]
Here, \(\operatorname{MSE}\), \(\operatorname{CE}\), and \(D_{\mathrm{KL}}\)
denote mean squared error, cross-entropy between the softmax distributions
of prediction logits, and Kullback--Leibler divergence, respectively.
Combining the model-component
losses gives
\[
\mathcal{L}_{\mathrm{model}}^{\mathrm{dist}}
=\frac{\beta_{\mathrm{dyn}}}{2}
\left(\mathcal{L}_h^{\mathrm{dist}}+\mathcal{L}_d^{\mathrm{dist}}\right)
+\beta_R\mathcal{L}_R^{\mathrm{dist}}
+\beta_Q\mathcal{L}_Q^{\mathrm{dist}}.
\]
The weights \(\beta_{\mathrm{dyn}},\beta_R,\beta_Q\) follow TD-MPC2's
model objective.

\subsection{Joint Multitask Adaptation}
\label{sec:method-objective}

For each new task, the two embeddings retrieved in
Sec.~\ref{sec:method-retrieval} initialize separate trainable embeddings for
the dynamics-related and task-related components. The agent then gathers
additional experience from the \(n\) new tasks and stores it in a replay
buffer \(\mathcal{B}_{\mathrm{new}}\). Collection and sampling are balanced
across tasks so that each contributes equally to new-task learning.

% Declare the experiment scene float here so it can occupy the top of page 5.
\begin{figure}[!t]
  \centering
  % Times New Roman glyph outlines avoid reader-dependent font substitution.
  \includegraphics[width=\linewidth]{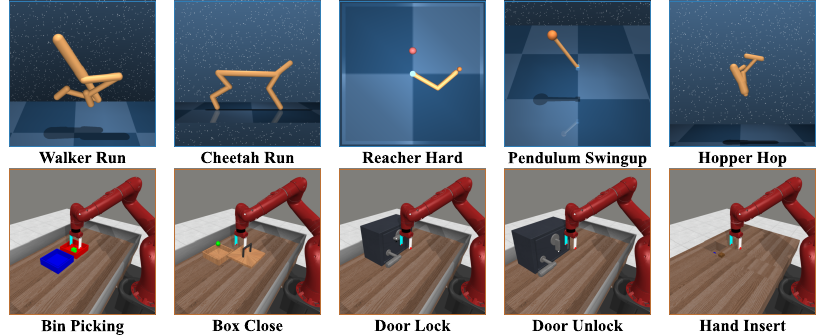}
  \caption{\textbf{The ten new tasks.} Five DeepMind Control tasks (top)
  and five Meta-World tasks (bottom).}
  \label{fig:new10-task-scenes}
\end{figure}

We use TD-MPC2's model and policy learning objectives, evaluated with each
new task's two embeddings. The model objective combines latent consistency,
reward prediction, and temporal-difference losses, while the policy objective
improves the policy prior using value estimates and entropy regularization.
Their averages over task-balanced batches from
\(\mathcal{B}_{\mathrm{new}}\) are denoted by
\(\mathcal{L}_{\mathrm{model}}^{\mathrm{new}}\) and
\(\mathcal{L}_{\mathrm{policy}}^{\mathrm{new}}\), respectively.

We combine new-task learning with capability retention through
\[
\begin{aligned}
\mathcal{L}_{\mathrm{model}}
&=\mathcal{L}_{\mathrm{model}}^{\mathrm{new}}
+\lambda_{\mathrm{model}}\mathcal{L}_{\mathrm{model}}^{\mathrm{dist}},\\
\mathcal{L}_{\mathrm{policy}}
&=\mathcal{L}_{\mathrm{policy}}^{\mathrm{new}}
+\lambda_{\mathrm{policy}}\mathcal{L}_\pi^{\mathrm{dist}}.
\end{aligned}
\]
Here, \(\lambda_{\mathrm{model}}\) and \(\lambda_{\mathrm{policy}}\) are
hyperparameters. Based on these objectives, we train a single agent to
acquire multiple new tasks simultaneously while retaining its previous
capabilities.

\subsection{Weight Adaptation with Fixed SVD Bases}
\label{sec:method-svd}

To optimize the joint multitask objectives in Sec.~\ref{sec:method-objective},
we introduce SVD-based weight adaptation as a complementary mechanism to
exemplar-based component distillation. While distillation aims to preserve
the functional behavior of the pretrained world model on old-task exemplars,
SVD-based adaptation expresses updates to the shared model weights within
the structured singular-vector bases learned during pretraining. Together,
these mechanisms seek to support new-task adaptation while retaining
previously acquired capabilities.

Following prior SVD-based adaptation methods~\citep{lingam2024svft,ge2026cersa},
we apply singular value decomposition to selected linear weight matrices of
the shared model. For each matrix \(W\), we can select \(r\) singular
directions and express its rank-\(r\) approximation and adapted form as
\[
W\approx U\Sigma V^\top,\qquad
W'=UCV^\top,\qquad
C_{\mathrm{init}}=\Sigma.
\]
Here, \(U\) and \(V\) contain the selected singular vectors, and \(\Sigma\)
is the diagonal matrix of corresponding singular values. We keep \(U\) and
\(V\) fixed and optimize the core matrix \(C\) under the joint objectives in
Sec.~\ref{sec:method-objective}. The core matrix can be diagonal, sparse,
or dense, allowing different degrees of flexibility in adjusting and
recombining pretrained directions. The suitability of these parameterizations
may depend on the richness and diversity of previously learned tasks and
their relatedness to newly introduced tasks.

\section{Experiments}
\label{sec:experiments}

We evaluate our framework on diverse continuous control tasks from the DeepMind Control
Suite~\citep{tassa2018dmcontrol} and Meta-World~\citep{yu2019metaworld}, spanning
classical control, locomotion, and robotic manipulation. We start from
TD-MPC2 models~\citep{hansen2024tdmpc2} trained on 70 tasks, using released
checkpoints where available, and adapt them to 10 new tasks simultaneously.
Figure~\ref{fig:new10-task-scenes} visualizes these ten new tasks.
The original 70 tasks are presented in Appendix~\ref{app:task-suite}.
Implementation details are provided in Appendix~\ref{app:algorithm-implementation}.

We seek to answer the following questions through experimentation:
\begin{itemize}
    \item Starting from a multitask world
    model trained on 70 tasks, can the representations retrieved in
    Sec.~\ref{sec:method-retrieval} provide an effective initialization for
    learning the 10 new tasks?

    \item Can our method
    accelerate learning of the 10 new tasks while largely preserving
    performance on the 70 previously learned tasks?

    \item How does each component introduced in
    Sec.~\ref{sec:online-multitask-incremental-learning} influence performance?

    \item How does our method scale with increasing model size?
\end{itemize}

% Keep the baseline heading with its introductory paragraph.
\Needspace{6\baselineskip}
\subsection{Baselines}
\label{sec:experiments-baselines}

We consider four baselines for evaluating new-task adaptation and retention
of existing capabilities. Three build on TD-MPC2~\citep{hansen2024tdmpc2}
and start from the same model trained on the 70 original tasks as MACR,
providing comparisons on the same backbone. A multitask Soft Actor-Critic
(Multi-SAC) baseline~\citep{haarnoja2018sac} provides an additional model-free
comparison. All methods adapt jointly to the 10 new tasks under the same
total environment-interaction budget.

(1) \textbf{TD-MPC2 FT with Experience Replay (ER)} initializes each new
task with a single retrieved task embedding shared across all components
and fine-tunes the model using the standard TD-MPC2 objectives. It also
replays stored transitions from the original tasks, applying the model
and policy learning objectives to these transitions to mitigate forgetting.
The original-task replay uses the same raw exemplar budget as MACR.

(2) \textbf{TD-MPC2 FT with Elastic Weight Consolidation (EWC)} uses the
same starting model and embedding initialization as the ER baseline.
EWC~\citep{kirkpatrick2017ewc} estimates parameter importance from
original-task exemplars using a diagonal Fisher approximation. During
fine-tuning, Fisher-weighted quadratic penalties discourage deviations of
the world-model and policy-prior parameters from their original values.

(3) \textbf{TD-MPC2 FT with ClonEx} adapts the policy-cloning
and best-return exploration mechanisms of
ClonEx-SAC~\citep{wolczyk2022clonex} to TD-MPC2's task-conditioned model
and MPC controller. For each new task, it evaluates compatible original-task
controllers and uses the highest-return controller to collect initial
exploration data. During joint adaptation, a KL loss on stored original-task
states aligns the current policy prior with the frozen original policy
prior. The remaining components are trained with the standard TD-MPC2
objectives.
% This is a TD-MPC2 adaptation; earlier Multi-SAC ClonEx results are distinct.

(4) \textbf{Multi-SAC with EWC} starts from a multitask SAC model trained
on the 70 original tasks. Each new task's actor and critic embeddings are
initialized from an existing task selected by minimizing the SAC
temporal-difference loss on a probing context. The model is then jointly
fine-tuned on the 10 new tasks using SAC objectives and EWC penalties on
parameters important for the original tasks.

\subsection{Results}
\label{sec:experiments-results}

\begin{figure}[!htbp]
  \noindent
  \begin{minipage}[t]{0.48\linewidth}
    \vspace{0pt}
    \centering
    % Font-free Times New Roman glyph outlines avoid cross-figure subset collisions.
    \includegraphics[width=\linewidth]{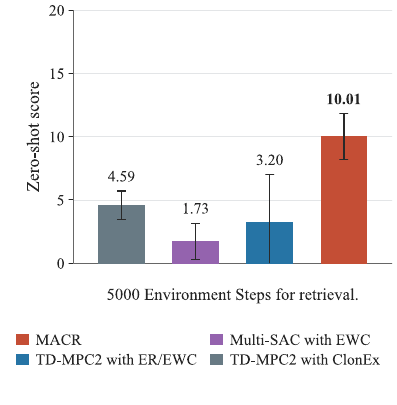}
  \end{minipage}\hfill
  \begin{minipage}[t]{0.48\linewidth}
    \vspace{0pt}
    \centering
    \includegraphics[width=\linewidth]{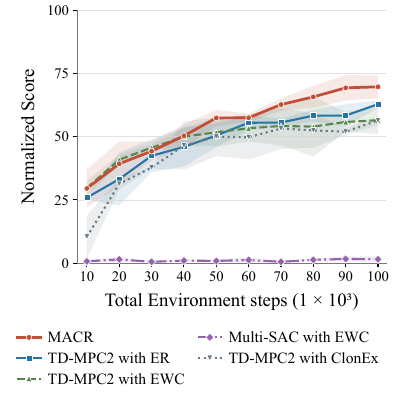}
  \end{minipage}
  \caption{\textbf{Zero shot and few-shot adaptation performance on the 10 new tasks.}
  \textbf{Left:} Zero shot performance on the 10 new tasks, with scores
  averaged over the 10 tasks.
  \textbf{Right:} Learning curves with 100k environment steps per task.
  Normalized scores are averaged over the 10 tasks.
  We report mean and standard deviation over three seeds}
  \label{fig:new10-5m-mean-learning}
  \label{fig:zero-shot-5m}
\end{figure}

\textbf{Zero shot performance.} MACR provides a stronger initialization for
adaptation. The left panel of Figure~\ref{fig:zero-shot-5m} evaluates each
method after its initial retrieval or controller-selection phase and before
any gradient update.
MACR achieves the highest zero-shot performance among all methods.
TD-MPC2 FT with ClonEx is the strongest baseline: executing the
best-performing compatible original-task controller already transfers part
of the required behavior. However, this controller is selected as a whole,
so its dynamics, reward, value, and policy behavior all come from a single
original task. MACR instead retrieves separate embeddings for the
dynamics-related and task-related components, allowing each to draw on the
original task whose behavior best matches it.
The ER and EWC baselines, which share the same TD-MPC2 starting point, and
Multi-SAC with EWC remain clearly below MACR, suggesting that their
initializations provide little directly usable behavior on the new tasks.
Table~\ref{tab:retrieved-embeddings} further shows that the selected embeddings
often come from tasks with related embodiments or behaviors.
For dynamics-related components, shared embodiments may lead to overlapping
state distributions, allowing the encoder to map new-task observations to
familiar latent regions and reuse learned dynamics. For task-related
components, similar reward structures and control behaviors may support
reuse of reward and value estimates to evaluate imagined trajectories,
while the retrieved policy prior guides action sampling during planning.
Appendix~\ref{app:retrieval-results} reports the retrieval results for all three seeds.

% Restricted retrieval, 5M, native run seed 3; no performance values in this table.
% Source: tdmpc2/tdmpc2/logs/mt70/3/
% split_world_policy_fullcore_gbd_5m_1m_seed3/split_retrieval_scores.json
% SHA256: 606f79a0a65d9768d54ce04cdea74eda442c06221169caa52db08aa0fd167d42
% Dynamics route: encoder h and dynamics d. Task route: reward R, value Q, policy pi.
% Human-readable names omit the mw- prefix; the domain headers identify those tasks.
% Allow the table to float near its introduction without leaving a page gap.
\begin{table}[!htbp]
  \centering
  \caption{\textbf{Retrieved original tasks' embeddings for the ten new tasks.}
  Each entry identifies the original task whose embedding is retrieved for a
  new task, separately for Dynamics-related and Task-related components.}
  \label{tab:retrieved-embeddings}
  \small
  \setlength{\tabcolsep}{2.5pt}
  \renewcommand{\arraystretch}{1.15}
  \resizebox{\linewidth}{!}{%
  \begin{tabular}{@{}l*{10}{c}@{}}
    \toprule
    & \multicolumn{5}{c}{\textbf{Meta-World}}
    & \multicolumn{5}{c}{\textbf{DeepMind Control}} \\
    \cmidrule(lr){2-6}\cmidrule(lr){7-11}
    \textbf{New task}
    & \shortstack{Bin\\Picking}
    & \shortstack{Box\\Close}
    & \shortstack{Door\\Lock}
    & \shortstack{Door\\Unlock}
    & \shortstack{Hand\\Insert}
    & \shortstack{Walker\\Run}
    & \shortstack{Cheetah\\Run}
    & \shortstack{Reacher\\Hard}
    & \shortstack{Pendulum\\Swingup}
    & \shortstack{Hopper\\Hop} \\
    \midrule
    Dynamics-related
    & \shortstack{Coffee\\Pull}
    & \shortstack{Pick\\Place\\Wall}
    & \shortstack{Basketball}
    & \shortstack{Pick\\Out of\\Hole}
    & \shortstack{Coffee\\Push}
    & \shortstack{Walker\\Walk\\Backwards}
    & \shortstack{Cheetah\\Run\\Front}
    & \shortstack{Reacher\\Easy}
    & \shortstack{Pendulum\\Spin}
    & \shortstack{Hopper\\Stand} \\
    \addlinespace[4pt]
    Task-related
    & \shortstack{Handle\\Press}
    & \shortstack{Faucet\\Open}
    & \shortstack{Drawer\\Open}
    & \shortstack{Window\\Close}
    & \shortstack{Push\\Back}
    & \shortstack{Walker\\Run\\Backwards}
    & \shortstack{Cheetah\\Run\\Front}
    & \shortstack{Reacher\\Easy}
    & \shortstack{Pendulum\\Spin}
    & \shortstack{Hopper\\Hop\\Backwards} \\
    \bottomrule
  \end{tabular}%
  }
\end{table}

% Source: results/mt70_source_scores_20260926/hopper_starting_scores.csv.
% Before-adaptation evaluations of the same starting models as Table 1.
\Needspace{14\baselineskip}
\begingroup
\setlength{\intextsep}{2pt}
\setlength{\columnsep}{6pt}
\begin{wraptable}{r}{0.42\linewidth}
  \centering
  \small
  \setlength{\abovecaptionskip}{0pt}
  \setlength{\belowcaptionskip}{2pt}
  \caption{\textbf{Hopper tasks' performance.} Normalized scores before
  adaptation, averaged over 10 episodes per task.}
  \label{tab:mt70-hopper-starting-scores}
  \setlength{\tabcolsep}{3pt}
  \begin{tabular*}{\linewidth}{@{\extracolsep{\fill}}lrr@{}}
    \toprule
    Task & TD-MPC2 & Multi-SAC \\
    \midrule
    Stand         & 0.47 & 0.12 \\
    Hop Backwards & 0.22 & 0.00 \\
    \bottomrule
  \end{tabular*}
\end{wraptable}

\textbf{Few-shot Adaptation.}
Figure~\ref{fig:new10-5m-mean-learning} (right) shows that MACR adapts
effectively to the new tasks. Appendix~\ref{app:new-task-learning-curves}
reports the learning curves for individual tasks. Among the TD-MPC2
baselines, EWC exhibits limited new-task learning, whereas ER learns
effectively. ClonEx also enables new-task learning, although its gains vary
across tasks; Multi-SAC with EWC makes limited progress. We attribute the
contrast between EWC and ER to their use of previously acquired knowledge:
EWC can restrict changes to shared parameters needed for new tasks, whereas
ER combines stored old-task transitions with new experience, providing
additional supervision for reusable representations and helping stabilize
world-model learning. For ClonEx, exploration transfer depends on the
behavioral compatibility between previously learned controllers and new
tasks. Policy cloning preserves old policy behavior but does not directly
provide a suitable initialization for the new task's dynamics, reward, and
value predictions. Behavioral transfer can therefore aid exploration
without providing equally effective initialization across all model
components. Hopper Hop remains challenging, consistent with weak
capabilities on the related original Hopper tasks
(Table~\ref{tab:mt70-hopper-starting-scores}).
\par
\endgroup

% Results scope: these figures establish initialization and New10 learning
% performance. Add post-adaptation Old70 results and component ablations
% before drawing retention or component-specific causal conclusions.
% Zero-update candidate sets and evaluation RNG schedules differ across
% methods; these are comparisons of full initialization procedures.

\paragraph{Retention on original tasks.}
We assess retention through the proportion of original-task performance
preserved after adaptation. TD-MPC2 with ER exhibits substantial forgetting
despite effective new-task learning, whereas EWC achieves retention
comparable to MACR but limits adaptation. In our problem setting, access to
large amounts of data from previously learned tasks is restricted. ER's TD
targets can bootstrap from value estimates at current-policy actions outside
the support of stored trajectories, while policy-improvement updates can
exploit these unconstrained estimates~\citep{fujimoto2019offpolicy,kumar2020conservative}.
These errors can directly impair TD-MPC2's action selection through its
learned reward and value functions. EWC instead protects parameters
important for old tasks, which may also be needed for new tasks sharing the
same model components. Its parameter-wise penalties do not distinguish
changes that disrupt old-task outputs from those that preserve them,
potentially restricting useful adaptation. MACR regularizes functional
behavior by aligning its outputs with those of the original model without
TD bootstrapping, and explicitly constrains the policy through a KL
divergence to the original policy. These functional constraints allow more
flexibility in parameter adaptation while preserving previously acquired
capabilities.
% Retention = 100 * mean Old70 score after / mean Old70 score before,
% computed per adaptation seed, then summarized across seeds (sample SD).
% Values supplied by the author on 2026-11-19, pending the author's check
% against source CSVs. Several differ from results/rebuttal_20261115/tables
% (MACR, ER, EWC); the evaluation budget is not stated.
% Keep this compact table on a page with the results text.
\begin{table}[!htb]
  \centering
  \caption{\textbf{Retention on the original 70 tasks.}
  Percentage of the pre-adaptation mean score retained after adaptation,
  using 5M models and 10 evaluation episodes per task. We report mean $\pm$
  sample standard deviation over three adaptation seeds.}
  \label{tab:old70-retention}
  \setlength{\tabcolsep}{10pt}
  \begin{tabular}{@{}lr@{}}
    \toprule
    Method & Retention (\%) \\
    \midrule
    TD-MPC2 + FT & $34.56 \pm 4.46$ \\
    TD-MPC2 + ER & $42.00 \pm 4.78$ \\
    TD-MPC2 + EWC & $86.00 \pm 1.33$ \\
    \textbf{MACR (Ours)} & $\mathbf{93.50 \pm 3.93}$ \\
    \bottomrule
  \end{tabular}
  % Trim the extra bottom gap of this float without changing the other tables.
  \par\vspace{-10pt}
\end{table}

\Needspace{9\baselineskip}
\paragraph{Scalability.}
\begingroup
\setlength{\intextsep}{2pt}
\setlength{\columnsep}{6pt}
% Source: results/model_scaling_20260926/summary.csv.
% World-and-policy FULLCORE with split embeddings; independent 1M evaluations.
\begin{wraptable}{r}{0.31\linewidth}
  \centering
  \small
  \setlength{\abovecaptionskip}{0pt}
  \setlength{\belowcaptionskip}{2pt}
  \caption{\textbf{Model scalability.} All-80 normalized scores.}
  \label{tab:model-scalability}
  \setlength{\tabcolsep}{7pt}
  \begin{tabular*}{\linewidth}{@{\extracolsep{\fill}}cc@{}}
    \toprule
    5M & 19M \\
    \midrule
    $45.10 \pm 1.80$ & $49.68 \pm 1.32$ \\
    \bottomrule
  \end{tabular*}
  \par\vspace{2pt}
  {\footnotesize mean $\pm$ std over 3 seeds}
\end{wraptable}
We increase the model size from 5M to 19M and evaluate the adapted agents
across all 80 tasks under the same 1M interaction budget. As shown in
Table~\ref{tab:model-scalability}, the larger model achieves higher overall
performance after adaptation with capability retention. This result suggests
that our method can benefit from increased model capacity, providing
preliminary evidence of its scalability.
\par
\endgroup

\paragraph{Ablation.}
To investigate the contributions of our method's main mechanisms to adaptation
and retention, we evaluate variants without exemplar-based distillation,
SVD-based parameterization, or embedding retrieval.
Table~\ref{tab:new10-component-ablations} reports performance on both the
70 original tasks and the 10 new tasks, including the retrieval-strategy
comparison described below.
The results indicate that exemplar-based distillation helps preserve
original-task performance when adapting coefficients within fixed SVD bases.
Embedding retrieval also benefits new-task learning, yielding higher average
performance on the new tasks.

% Source: results/new10_ablation_1m_20260926/summary.csv.
% Independent final-checkpoint evaluation, not periodic learning-curve endpoints.
% Retrieval-strategy scores and retention percentages: collaborator-provided
% results/split_retrieval_ablation_20261119/summary.csv; zero-shot results omitted.
\begin{table}[!htbp]
  \centering
  \caption{\textbf{Component and retrieval ablations.}
  We report normalized scores.}
  \label{tab:new10-component-ablations}
  \small
  \setlength{\tabcolsep}{6pt}
  \begin{tabular}{@{}lrr@{}}
    \toprule
    Method & \shortstack{Pretrained 70\\tasks} & \shortstack{New 10\\tasks} \\
    \midrule
    \textbf{MACR (Ours)} & $\mathbf{43.45 \pm 2.41}$ & $\mathbf{73.41 \pm 6.88}$ \\
    w/o Retrieval & $38.02 \pm 2.31$ & $67.35 \pm 6.60$ \\
    w/o Exemplar Distillation & $17.11 \pm 1.14$ & $68.73 \pm 3.97$ \\
    w/o SVD Decomposition & $28.33 \pm 0.83$ & $72.91 \pm 4.57$ \\
    Dynamics-only retrieval & $41.64 \pm 2.63$ & $70.64 \pm 2.47$ \\
    Task-related-only retrieval & $39.18 \pm 1.33$ & $65.82 \pm 4.51$ \\
    \bottomrule
  \end{tabular}
\end{table}

We further isolate the contribution of split embedding retrieval by comparing
three retrieval strategies while keeping all other adaptation settings,
including SVD-based adaptation and exemplar-based distillation, unchanged.
\textbf{Dynamics-only retrieval} selects a single embedding using
\(\mathcal{L}_{\mathrm{dyn}}\) and shares it across all model components.
\textbf{Task-related-only retrieval} instead selects a single shared embedding
using \(\beta_R\mathcal{L}_R+\beta_Q\mathcal{L}_Q\).
\textbf{Split retrieval (MACR)} retrieves separate embeddings for the
dynamics-related and task-related components using their respective
objectives, as described in Sec.~\ref{sec:method-retrieval}.
Table~\ref{tab:new10-component-ablations} shows that dynamics-only retrieval
achieves higher mean new-task performance than task-related-only retrieval,
while split retrieval achieves the highest mean new-task performance.
This pattern is consistent with the model-based planning mechanism:
accurate dynamics predictions support model-based planning, reward and value
estimates guide trajectory evaluation, and the actor, trained using value
estimates, proposes action candidates. These results support separately
retrieving dynamics-related and task-related representations to reuse both
aspects of the pretrained world model for new-task adaptation.

\section{Conclusion}
\label{sec:conclusion}

We presented MACR, a framework for adapting trained multitask world models
to multiple new continuous control tasks while retaining prior capabilities.
Through brief exploration of each new task, MACR retrieves dynamics-related
and task-related representations to provide an effective initialization for
adaptation. Adaptation combines learning core weight matrices within fixed
SVD bases with exemplar-based distillation to balance new-task learning and
knowledge retention. Our exploratory experiments on DMControl and Meta-World
show improved new-task learning over a range of adaptation baselines while
retaining most of the original-task performance. These findings suggest that
reusing and preserving knowledge in existing multitask world models offers a
promising route to advancing generalist agents as new tasks become accessible.

% TODO FINAL: Acknowledgements are omitted from the anonymous review
% submission and should only be added for the camera-ready version, e.g.:
% \section*{Acknowledgements}

\subsection*{AI Use Statement}
% Working disclosure based on documented use as of 2026-09-24.
% Before submission, the authors should add any other tools/uses and describe
% the human checks they actually completed. Do not claim comprehensive human
% review or non-use for unverified categories. Mirror the scope in the
% submission form. Policy: https://iclr.cc/Conferences/2027/AIPolicyForAuthors
We used OpenAI Codex to assist with refining the problem formulation and
experimental design, implementing and debugging methods, baselines, and
evaluation scripts. It also assisted with literature search, translating and revising
manuscript text. The authors direct the research and take responsibility for the final manuscript, implementations, experimental
results, and scientific claims, including material produced with AI assistance.

\bibliography{main}
\bibliographystyle{iclr2027_conference}
\clearpage
\appendix
% The official ICLR style suppresses \addcontentsline. Keep an appendix-only
% contents list without changing the style; names, numbers, and pages all
% resolve from labels, and each complete entry links to its destination.
\section*{Appendix Contents}
\begingroup
\setlength{\parskip}{2pt}
\newcommand{\appendixentry}[2][0pt]{%
  \par\noindent\hspace*{#1}%
  \hyperref[#2]{\ref*{#2}\quad\nameref*{#2}\dotfill\pageref*{#2}}%
}
\appendixentry{app:algorithm-implementation}
\appendixentry[1.5em]{app:algorithm}
\appendixentry[1.5em]{app:interaction-memory}
\appendixentry[1.5em]{app:svd-bases}
\appendixentry{app:task-suite}
\appendixentry[1.5em]{app:task-suite-dmcontrol}
\appendixentry[1.5em]{app:task-suite-metaworld}
\appendixentry{app:retrieval-results}
\appendixentry{app:taskwise-results}
\appendixentry[1.5em]{app:new-task-learning-curves}
\appendixentry[1.5em]{app:taskwise-retention}
\par
\endgroup
\medskip

\section{Algorithm and Implementation Details}
\label{app:algorithm-implementation}
\subsection{Algorithm Overview}
\label{app:algorithm}

Algorithm~\ref{alg:macr} summarizes the TD-MPC2 instantiation of MACR.

% Core LaTeX only: keep the pseudocode usable with the cached paper toolchain.
\newcounter{macralgorithm}
\newcounter{macrline}
\noindent
\begin{minipage}{\linewidth}
\small
\refstepcounter{macralgorithm}\label{alg:macr}
\hrule
\vspace{4pt}
\textbf{Algorithm \themacralgorithm\quad MACR: retrieval, adaptation, and retention}
\par\vspace{4pt}\hrule\vspace{5pt}
\textbf{Input:} Original model $\bar f=f(\theta_0,E_0)$; exemplar memory
$\mathcal B_{\mathrm{pre}}$; new tasks $\mathcal T_{\mathrm{new}}$;
candidate sets $\mathcal E_i\subseteq\mathcal E$; interaction budget.
\begin{list}{\arabic{macrline}:}{%
  \usecounter{macrline}%
  \setlength{\leftmargin}{1.8em}%
  \setlength{\labelwidth}{1.4em}%
  \setlength{\labelsep}{.4em}%
  \setlength{\itemsep}{1pt}%
  \setlength{\parsep}{0pt}%
  \setlength{\topsep}{4pt}%
}
  \item Initialize the adapted model from $\bar f$; keep $\bar f$ and $E_0$ fixed.
  \item For each adapted weight $W=U\Sigma V^\top$, set $W'=UCV^\top$, $C\gets\Sigma$.
  \item Train dense cores and task embeddings; freeze bases and other parameters.
  \item Cache the original component outputs on $\mathcal B_{\mathrm{pre}}$ as fixed targets.
  \item Initialize new-task replay $\mathcal B_{\mathrm{new}}\gets\emptyset$.
  \item \textbf{for each} new task $\mathcal M_i\in\mathcal T_{\mathrm{new}}$ \textbf{do}
  \item \hspace*{1em}Collect random-action context $\mathcal C_i$ within the interaction budget.
  \item \hspace*{1em}$e_i^{\mathrm{dyn}}\gets
    \arg\min_{e_j\in\mathcal E_i}\mathcal L_{\mathrm{dyn}}(e_j;\mathcal C_i)$.
  \item \hspace*{1em}Hold $e_i^{\mathrm{dyn}}$ fixed for $(h,d)$ while scoring control candidates.
  \item \hspace*{1em}$e_i^{\mathrm{ctrl}}\gets
    \arg\min_{e_j\in\mathcal E_i}
    [\beta_R\mathcal L_R(e_j;\mathcal C_i)+\beta_Q\mathcal L_Q(e_j;\mathcal C_i)]$.
  \item \hspace*{1em}Copy these into trainable rows for $(h,d)$ and $(R,Q,\pi)$, respectively.
  \item \hspace*{1em}Add $\mathcal C_i$ to $\mathcal B_{\mathrm{new}}$.
  \item \textbf{end for}
  \item \textbf{for each} scheduled learning update \textbf{do}
  \item \hspace*{1em}When scheduled, collect equally from all new tasks using the current planner;
    append this experience to $\mathcal B_{\mathrm{new}}$ within the remaining budget.
  \item \hspace*{1em}Sample task-balanced new-task and exemplar batches from their buffers.
  \item \hspace*{1em}Compute TD-MPC2 losses $\mathcal L_{\mathrm{model}}^{\mathrm{new}}$,
    $\mathcal L_{\mathrm{policy}}^{\mathrm{new}}$ on the new-task batch.
  \item \hspace*{1em}Compute $\mathcal L_{\mathrm{model}}^{\mathrm{dist}}$ and
    $\mathcal L_\pi^{\mathrm{dist}}$ from fixed exemplar targets
    (Sec.~\ref{sec:method-distillation}).
  \item \hspace*{1em}$\mathcal L_{\mathrm{model}}\gets
    \mathcal L_{\mathrm{model}}^{\mathrm{new}}
    +\lambda_{\mathrm{model}}\mathcal L_{\mathrm{model}}^{\mathrm{dist}}$.
  \item \hspace*{1em}$\mathcal L_{\mathrm{policy}}\gets
    \mathcal L_{\mathrm{policy}}^{\mathrm{new}}
    +\lambda_{\mathrm{policy}}\mathcal L_\pi^{\mathrm{dist}}$.
  \item \hspace*{1em}Update world-model cores and all old/new embeddings using $\mathcal L_{\mathrm{model}}$.
  \item \hspace*{1em}Update policy cores using $\mathcal L_{\mathrm{policy}}$; update the target value network.
  \item \textbf{end for}
  \item \textbf{return} The adapted model for $\mathcal T_{\mathrm{pre}}\cup\mathcal T_{\mathrm{new}}$.
\end{list}
\vspace{2pt}\hrule
\end{minipage}

% All three split_world_policy_fullcore_gbd_5m_1m_seed{1,2,3} manifests record
% candidate_rule="same domain and action dimension". The trainer implements
% this as name.split("-", 1)[0] equality plus action-dimension equality:
% Walker tasks share "walker"; all Meta-World tasks share "mw".
The three-seed 5M adaptation experiments restrict $\mathcal E_i$ to original
tasks with the same environment family and action dimension (e.g., Walker
tasks); Meta-World tasks are grouped together.

\clearpage
\subsection{Algorithm Implementations}
\label{app:interaction-memory}
\label{sec:experiments-implementations}

MACR collects 500 random-action transitions per new task as exploration.
Each candidate embedding is scored on the same context as described in
Sec.~\ref{sec:method-retrieval}; context collection counts toward the
adaptation budget. Multi-SAC with ClonEx receives the same budget of 500 environment
interactions per task in the initialization comparison. It evaluates the
70 policies learned on the original tasks and selects the highest-return
policy as the initial behavior policy for the new task.
The memory \(\mathcal{B}_{\mathrm{pre}}\) retains the four highest-return
trajectories from each original task. These trajectories, paired with the
corresponding outputs of each component of the original model, form the
exemplars. For distillation, we balance exemplar sampling across tasks.
For each adapted weight matrix
\(W_\ell\in\mathbb{R}^{d_{\mathrm{out}}\times d_{\mathrm{in}}}\), we
retain all \(r_\ell=\min(d_{\mathrm{out}},d_{\mathrm{in}})\) singular
directions, preserving all singular-value energy. We freeze the left and
right bases and optimize each dense
\(r_\ell\times r_\ell\) core, initialized to the diagonal
singular-value matrix. Further details are provided in
Table~\ref{tab:interaction-memory-counts}.

\begin{table}[htbp]
  \centering
  \setlength{\tabcolsep}{4pt}
  \caption{\textbf{Interaction and exemplar counts.} Totals cover the ten
  new tasks for interaction budgets and the 70 original tasks for memory.
  Domain-specific rows cover the five new tasks in that domain.}
  \label{tab:interaction-memory-counts}
  \small
  \begin{tabular}{@{}p{.42\linewidth}p{.29\linewidth}p{.23\linewidth}@{}}
    \toprule
    Quantity & Per task & Total \\
    \midrule
    \multicolumn{3}{@{}l}{\textit{MACR retrieval}} \\
    DMC random-action trajectories & $1\times500$ steps & 5 trajectories \\
    Meta-World random-action trajectories & $5\times100$ steps & 25 trajectories \\
    Collected context & 500 interactions & 5,000; 30 trajectories \\
    \midrule
    \multicolumn{3}{@{}l}{\textit{ClonEx initialization comparison}} \\
    Selection budget cap & 500 interactions & 5,000 interactions \\
    Actual policy trials & $70\times7$ steps & 4,900 interactions \\
    \midrule
    \multicolumn{3}{@{}l}{\textit{ClonEx adaptation curves}} \\
    Full-episode trials, DMC & $70\times500$ steps & 175,000 interactions \\
    Full-episode trials, Meta-World & $70\times100$ steps & 35,000 interactions \\
    Subsequent best-policy exploration & 10,000 steps & 100,000 interactions \\
    Updates begin at & 1,000 interactions & --- \\
    \midrule
    \multicolumn{3}{@{}l}{\textit{Exemplar memory}} \\
    Retained trajectory records & $4\times100$ transitions & 280 records \\
    Retained transitions & 400 & 28,000 \\
    ClonEx pre-transition observations & 400 & 28,000 \\
    MACR distillation window & 3 transitions & --- \\
    \bottomrule
  \end{tabular}
\end{table}

\Needspace{19\baselineskip}
\subsection{SVD Basis Counts}
\label{app:svd-bases}
\begin{table}[htbp]
  \centering
  \caption{SVD basis counts for the 5M model. Retained singular directions
  are listed in layer order. The Q row repeats the three listed ranks for
  each of five heads; its matrix and direction counts include all five.
  Each retained direction corresponds to one left--right singular-vector pair.}
  \label{tab:svd-basis-counts}
  \begin{tabular}{lrrr}
    \toprule
    Component & Matrices & Directions per matrix & Total pairs \\
    \midrule
    Encoder & 2 & $(135,256)$ & 391 \\
    Dynamics & 3 & $(512,512,512)$ & 1,536 \\
    Reward predictor & 3 & $(512,512,101)$ & 1,125 \\
    Policy prior & 3 & $(512,512,12)$ & 1,036 \\
    Q ensemble (5 heads) & 15 & $(512,512,101)$ per head & 5,625 \\
    \midrule
    Total & 26 & --- & 9,713 \\
    \bottomrule
  \end{tabular}
\end{table}

The dense $r_\ell\times r_\ell$ cores contain
4,601,559 trainable coefficients in total; task embeddings are not
SVD-factorized.

\clearpage
\section{Original Task Suite}
\label{app:task-suite}
\begin{figure}[!ht]
  \centering
  \includegraphics[width=\linewidth]{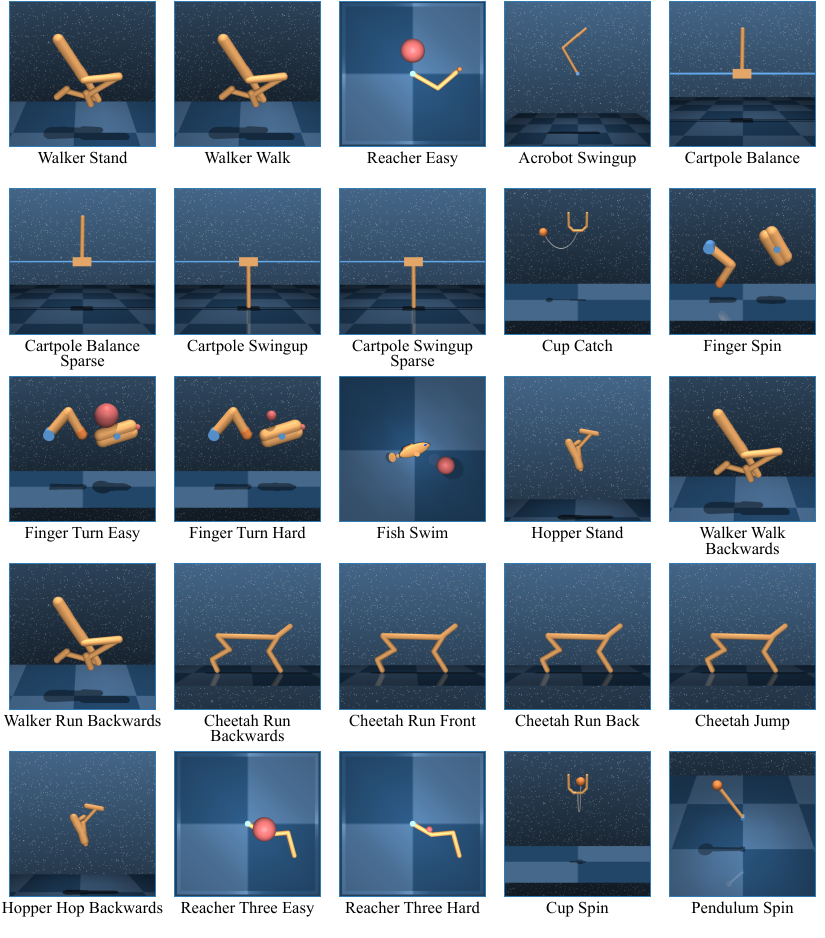}
  \caption{\textbf{Original DeepMind Control tasks.} The 25 tasks included
  in the original 70-task training set.}
  \label{fig:old70-dmcontrol}
\end{figure}

\begin{figure}[!ht]
  \centering
  \includegraphics[width=\linewidth]{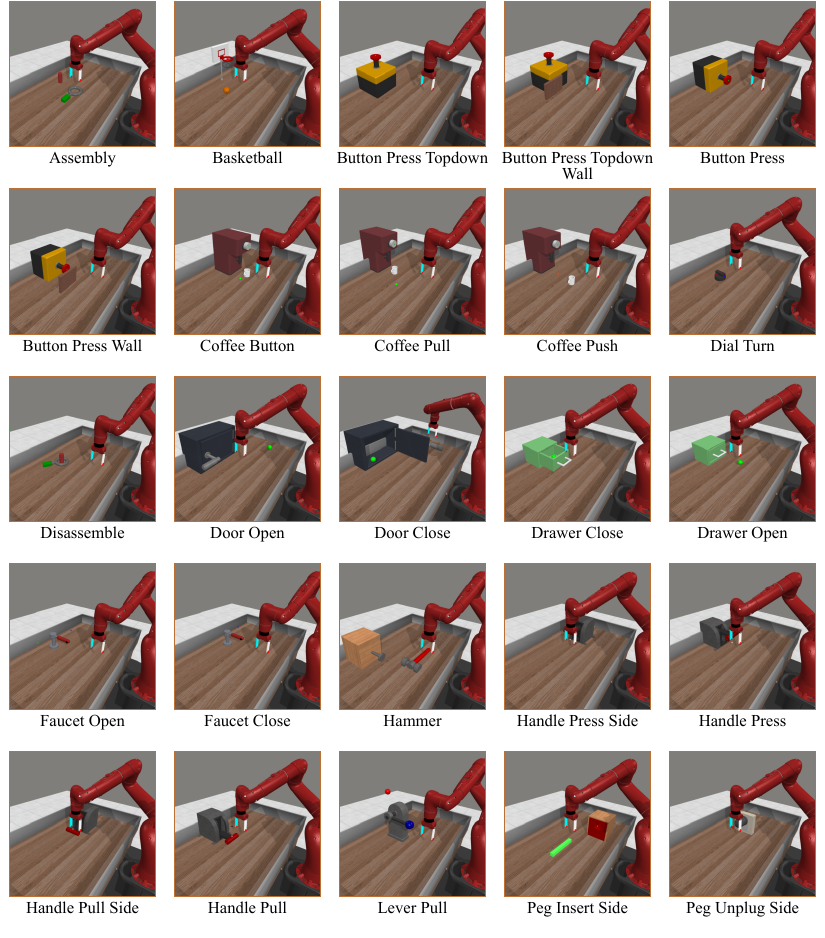}
  \caption{\textbf{Original Meta-World tasks, part 1.} The first 25 of
  the 45 manipulation tasks included in the original training set.}
  \label{fig:old70-metaworld}
\end{figure}

\begin{figure}[!t]
  \centering
  \includegraphics[width=\linewidth]{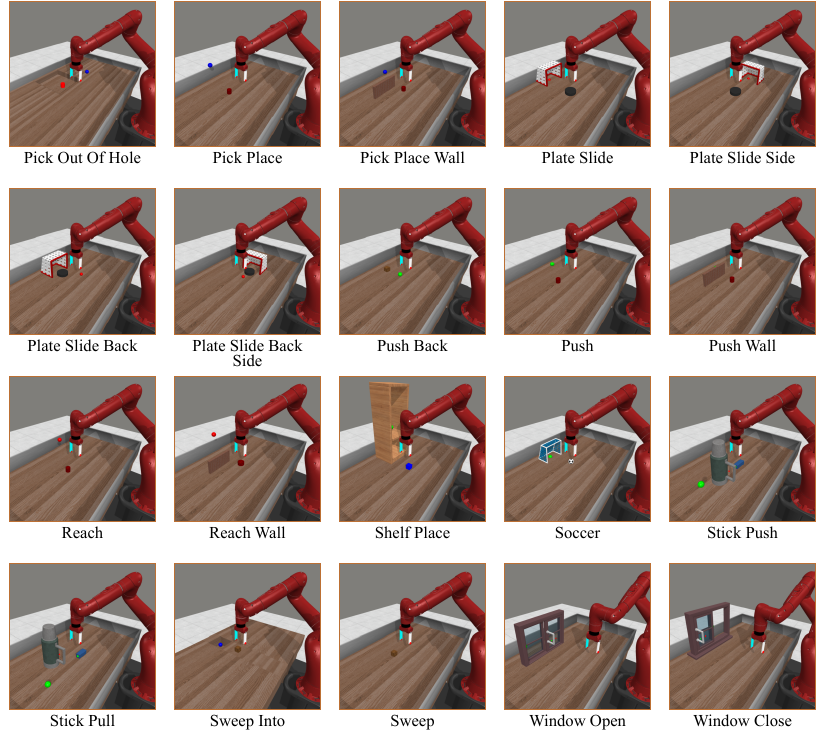}
  \caption{\textbf{Original Meta-World tasks, part 2.} The remaining 20
  manipulation tasks included in the original training set.}
  \label{fig:old70-metaworld-continued}
\end{figure}

The original task set contains 25 DeepMind Control tasks and 45 Meta-World
tasks. Figures~\ref{fig:old70-dmcontrol}--\ref{fig:old70-metaworld-continued}
show all 70 tasks and their names. These tasks are disjoint from the ten
new tasks in Figure~\ref{fig:new10-task-scenes}; together, the two sets
contain 30 DeepMind Control tasks and 50 Meta-World tasks.

\subsection{DeepMind Control Tasks}
\label{app:task-suite-dmcontrol}

\clearpage
\subsection{Meta-World Tasks}
\label{app:task-suite-metaworld}

\clearpage

\clearpage
\section{Embedding Retrieval and Task Representations}
\label{app:retrieval-results}
\paragraph{Sensitivity to random seeds.}
Random probing contexts and stochastic candidate scoring can change the
retrieval ranking. We repeat retrieval with seeds 1--3, keeping the trained
5M model and the budget of 500 transitions per new task fixed, to assess
the sensitivity of the selected embeddings and their initial performance.

Tables~\ref{tab:retrieval-initial-performance}
and~\ref{tab:retrieved-embeddings-three-seeds} report zero-update performance
and the selected source tasks. Dynamics-related choices agree across all
three seeds for nine of ten tasks, compared with six for task-related
choices; all disagreements occur on Meta-World tasks. Score variability
includes both retrieval and evaluation randomness.

\begin{table}[!htbp]
  \centering
  \caption{\textbf{Initial performance with retrieved embeddings.}
  Mean normalized score across the ten new tasks, before gradient updates.}
  \label{tab:retrieval-initial-performance}
  \setlength{\tabcolsep}{12pt}
  \begin{tabular}{@{}cccc@{}}
    \toprule
    Seed 1 & Seed 2 & Seed 3 & Mean $\pm$ SD \\
    \midrule
    11.03 & 7.93 & 11.07 & $10.01 \pm 1.80$ \\
    \bottomrule
  \end{tabular}
\end{table}

% Generated table only; explanatory prose is in appendix_retrieval_results.tex.
\begin{table}[!htbp]
  \centering
  \caption{\textbf{Retrieved original-task embeddings across three seeds.}
  Dynamics-related and Task-related selections for the ten new tasks
  using the 5M model (adaptation seeds 1--3).}
  \label{tab:retrieved-embeddings-three-seeds}
  \small
  \setlength{\tabcolsep}{2.5pt}
  \renewcommand{\arraystretch}{1.15}
  \textbf{Seed 1}\par\smallskip
  \resizebox{\linewidth}{!}{%
  \begin{tabular}{@{}l*{10}{c}@{}}
    \toprule
    & \multicolumn{5}{c}{\textbf{Meta-World}}
    & \multicolumn{5}{c}{\textbf{DeepMind Control}} \\
    \cmidrule(lr){2-6}\cmidrule(lr){7-11}
    \textbf{New task}
    & \shortstack{Bin\\Picking}
    & \shortstack{Box\\Close}
    & \shortstack{Door\\Lock}
    & \shortstack{Door\\Unlock}
    & \shortstack{Hand\\Insert}
    & \shortstack{Walker\\Run}
    & \shortstack{Cheetah\\Run}
    & \shortstack{Reacher\\Hard}
    & \shortstack{Pendulum\\Swingup}
    & \shortstack{Hopper\\Hop} \\
    \midrule
    Dynamics-related
    & \shortstack{Coffee\\Push}
    & \shortstack{Pick\\Place\\Wall}
    & \shortstack{Basketball}
    & \shortstack{Pick\\Out of\\Hole}
    & \shortstack{Coffee\\Push}
    & \shortstack{Walker\\Walk\\Backwards}
    & \shortstack{Cheetah\\Run\\Front}
    & \shortstack{Reacher\\Easy}
    & \shortstack{Pendulum\\Spin}
    & \shortstack{Hopper\\Stand} \\
    \addlinespace[4pt]
    Task-related
    & \shortstack{Stick\\Push}
    & \shortstack{Reach\\Wall}
    & \shortstack{Drawer\\Open}
    & \shortstack{Window\\Close}
    & \shortstack{Push\\Back}
    & \shortstack{Walker\\Run\\Backwards}
    & \shortstack{Cheetah\\Run\\Front}
    & \shortstack{Reacher\\Easy}
    & \shortstack{Pendulum\\Spin}
    & \shortstack{Hopper\\Hop\\Backwards} \\
    \bottomrule
  \end{tabular}%
  }
  \par\medskip
  \textbf{Seed 2}\par\smallskip
  \resizebox{\linewidth}{!}{%
  \begin{tabular}{@{}l*{10}{c}@{}}
    \toprule
    & \multicolumn{5}{c}{\textbf{Meta-World}}
    & \multicolumn{5}{c}{\textbf{DeepMind Control}} \\
    \cmidrule(lr){2-6}\cmidrule(lr){7-11}
    \textbf{New task}
    & \shortstack{Bin\\Picking}
    & \shortstack{Box\\Close}
    & \shortstack{Door\\Lock}
    & \shortstack{Door\\Unlock}
    & \shortstack{Hand\\Insert}
    & \shortstack{Walker\\Run}
    & \shortstack{Cheetah\\Run}
    & \shortstack{Reacher\\Hard}
    & \shortstack{Pendulum\\Swingup}
    & \shortstack{Hopper\\Hop} \\
    \midrule
    Dynamics-related
    & \shortstack{Coffee\\Pull}
    & \shortstack{Pick\\Place\\Wall}
    & \shortstack{Basketball}
    & \shortstack{Pick\\Out of\\Hole}
    & \shortstack{Coffee\\Push}
    & \shortstack{Walker\\Walk\\Backwards}
    & \shortstack{Cheetah\\Run\\Front}
    & \shortstack{Reacher\\Easy}
    & \shortstack{Pendulum\\Spin}
    & \shortstack{Hopper\\Stand} \\
    \addlinespace[4pt]
    Task-related
    & \shortstack{Button\\Press\\Wall}
    & \shortstack{Button\\Press\\Topdown\\Wall}
    & \shortstack{Window\\Close}
    & \shortstack{Button\\Press\\Topdown\\Wall}
    & \shortstack{Push\\Back}
    & \shortstack{Walker\\Run\\Backwards}
    & \shortstack{Cheetah\\Run\\Front}
    & \shortstack{Reacher\\Easy}
    & \shortstack{Pendulum\\Spin}
    & \shortstack{Hopper\\Hop\\Backwards} \\
    \bottomrule
  \end{tabular}%
  }
  \par\medskip
  \textbf{Seed 3}\par\smallskip
  \resizebox{\linewidth}{!}{%
  \begin{tabular}{@{}l*{10}{c}@{}}
    \toprule
    & \multicolumn{5}{c}{\textbf{Meta-World}}
    & \multicolumn{5}{c}{\textbf{DeepMind Control}} \\
    \cmidrule(lr){2-6}\cmidrule(lr){7-11}
    \textbf{New task}
    & \shortstack{Bin\\Picking}
    & \shortstack{Box\\Close}
    & \shortstack{Door\\Lock}
    & \shortstack{Door\\Unlock}
    & \shortstack{Hand\\Insert}
    & \shortstack{Walker\\Run}
    & \shortstack{Cheetah\\Run}
    & \shortstack{Reacher\\Hard}
    & \shortstack{Pendulum\\Swingup}
    & \shortstack{Hopper\\Hop} \\
    \midrule
    Dynamics-related
    & \shortstack{Coffee\\Pull}
    & \shortstack{Pick\\Place\\Wall}
    & \shortstack{Basketball}
    & \shortstack{Pick\\Out of\\Hole}
    & \shortstack{Coffee\\Push}
    & \shortstack{Walker\\Walk\\Backwards}
    & \shortstack{Cheetah\\Run\\Front}
    & \shortstack{Reacher\\Easy}
    & \shortstack{Pendulum\\Spin}
    & \shortstack{Hopper\\Stand} \\
    \addlinespace[4pt]
    Task-related
    & \shortstack{Handle\\Press}
    & \shortstack{Faucet\\Open}
    & \shortstack{Drawer\\Open}
    & \shortstack{Window\\Close}
    & \shortstack{Push\\Back}
    & \shortstack{Walker\\Run\\Backwards}
    & \shortstack{Cheetah\\Run\\Front}
    & \shortstack{Reacher\\Easy}
    & \shortstack{Pendulum\\Spin}
    & \shortstack{Hopper\\Hop\\Backwards} \\
    \bottomrule
  \end{tabular}%
  }
\end{table}

\clearpage
\paragraph{Effect of the probing context.}
To isolate variation due to the probing data, we also evaluate a fixed
19M model with three independently sampled 500-transition contexts per task.
Candidate-scoring randomness and the ten evaluation episodes are held fixed
across contexts. Table~\ref{tab:retrieval-context-sensitivity} compares
retrieval within the same environment family and action dimension with
retrieval over all 70 original tasks. The variation across contexts shows
why a single probing seed is insufficient to characterize initialization.

\begin{table}[!htbp]
  \centering
  \caption{\textbf{Retrieval performance across probing contexts.}
  Mean normalized score on the ten new tasks before gradient updates
  with the 19M model. SD is across the three context seeds.}
  \label{tab:retrieval-context-sensitivity}
  \setlength{\tabcolsep}{7pt}
  \begin{tabular}{@{}lcccc@{}}
    \toprule
    Candidate pool & Context 1 & Context 2 & Context 3 & Mean $\pm$ SD \\
    \midrule
    Same family and action dimension & 15.17 & 6.17 & 8.22 & $9.86 \pm 4.72$ \\
    All 70 original tasks & 15.48 & 6.49 & 14.53 & $12.17 \pm 4.93$ \\
    \bottomrule
  \end{tabular}
\end{table}

\begin{figure}[!htbp]
  \centering
  \includegraphics[width=\linewidth]{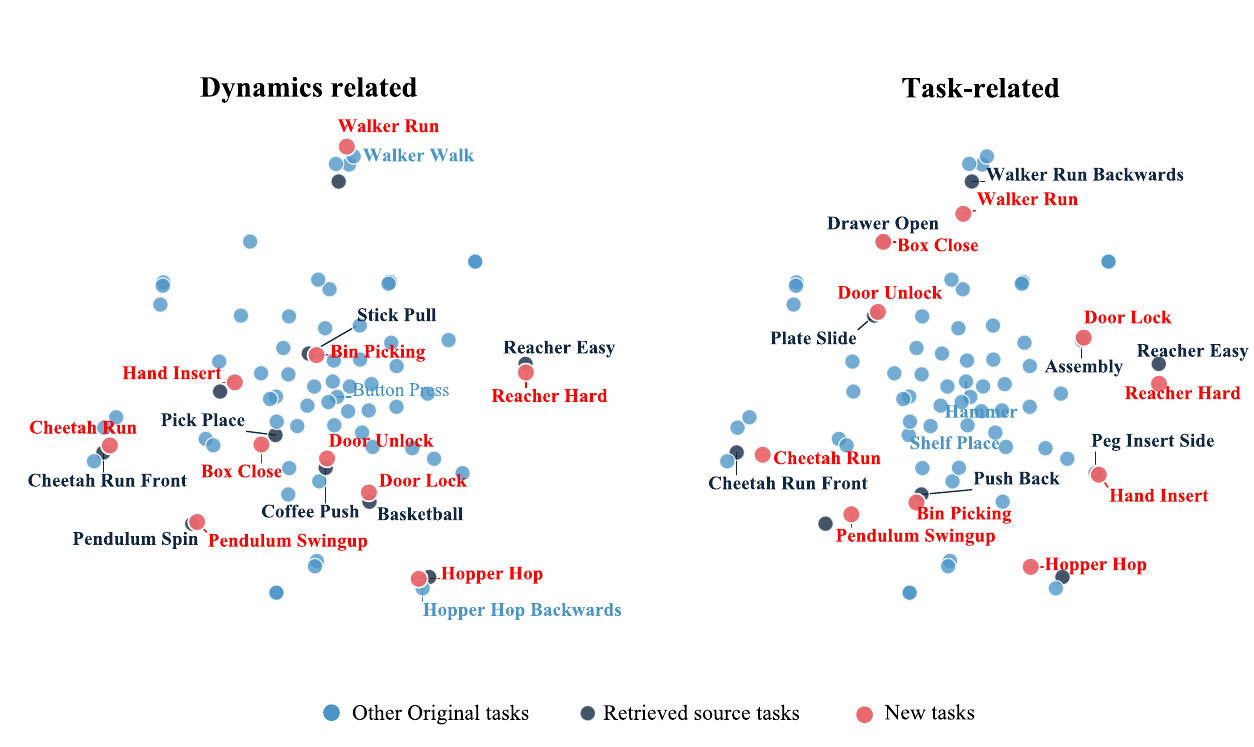}
  \caption{\textbf{Task embeddings after adaptation.}
  Dynamics-related and task-related views of one t-SNE projection of the
  96-dimensional embeddings. Red denotes new tasks; dark blue marks original
  tasks selected for retrieval initialization; light blue marks other
  original tasks.}
  \label{fig:adapted-task-embeddings}
\end{figure}

\paragraph{Task representations after adaptation.}
Figure~\ref{fig:adapted-task-embeddings} visualizes the task embeddings of
the 19M model after 0.5M total new-task interactions (seed 1).
A single t-SNE projection contains the 70 original-task vectors and the
two vectors for each of the ten new tasks. The panels share the original-task
coordinates and show the corresponding new dynamics-related or task-related
vectors. For example, Bin Picking lies near Stick Pull in the dynamics view
and Push Back in the task-related view, illustrating the different task
relationships represented by the two component groups.

\clearpage
\section{Task-wise Adaptation and Retention}
\label{app:taskwise-results}
\subsection{Learning Curves on New Tasks}
\label{app:new-task-learning-curves}
\begin{figure}[!ht]
  \centering
  \includegraphics[width=\linewidth]{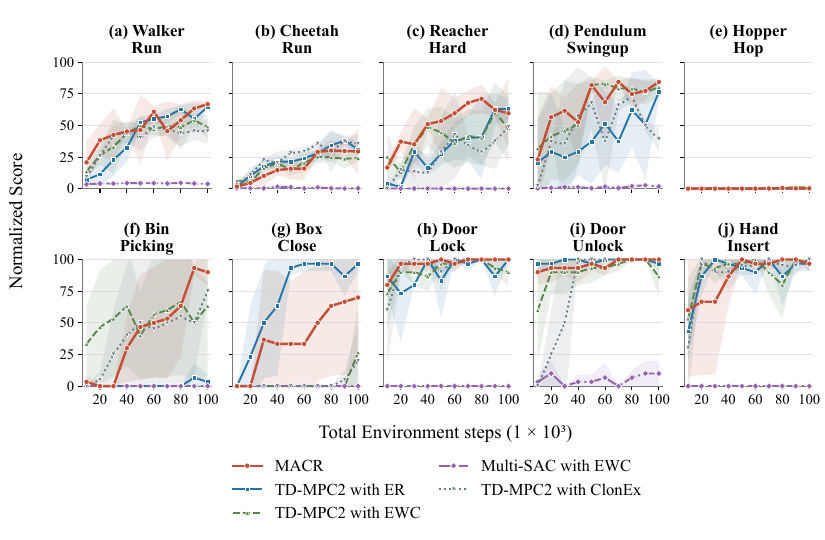}
  \caption{\textbf{Learning curves on the 10 new tasks.}
  Each task is allowed at most 100k environment interactions.
  We report the mean and standard deviation over two seeds for TD-MPC2
  with ClonEx and three seeds for all other methods.}
  \label{fig:new10-5m-learning-curves}
\end{figure}

\clearpage
\subsection{Retention on Original Tasks}
\label{app:taskwise-retention}
\begin{figure}[!ht]
  \centering
  \includegraphics[width=\linewidth]{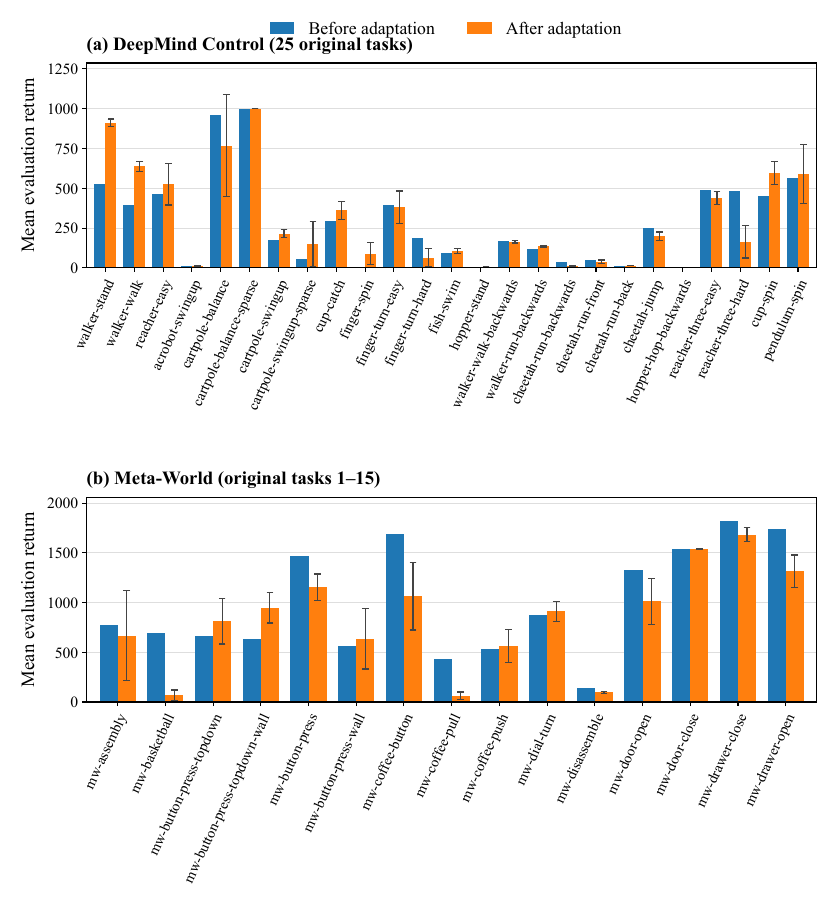}
  \caption{\textbf{Original-task performance before and after adaptation, part 1.}
  All 25 DeepMind Control tasks and the first 15 Meta-World tasks.
  Blue: the shared original-model reference, evaluated once without seed
  error bars. Orange: mean and sample standard deviation over three
  adaptation seeds. Each evaluation uses 10 episodes per task, separately
  from adaptation.}
  \label{fig:old70-before-after-part1}
\end{figure}

\begin{figure}[!t]
  \centering
  \includegraphics[width=\linewidth]{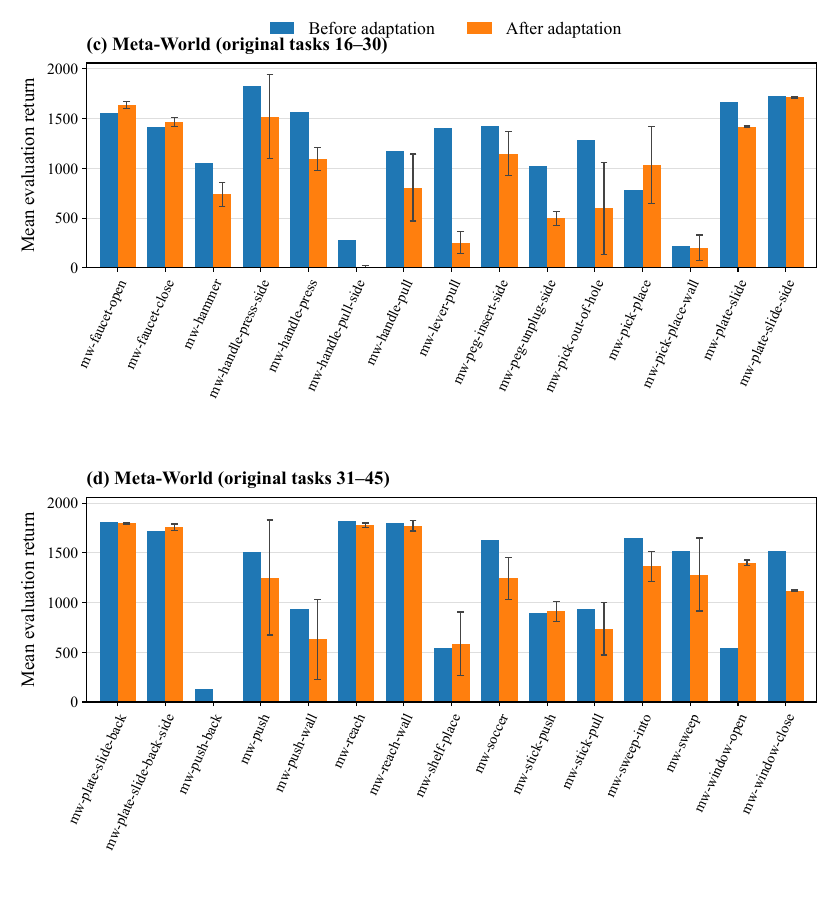}
  \caption{\textbf{Original-task performance before and after adaptation, part 2.}
  The remaining 30 Meta-World tasks, using the same protocol, colors, and
  uncertainty convention as Figure~\ref{fig:old70-before-after-part1}.
  All three Meta-World panels use the same vertical scale.}
  \label{fig:old70-before-after-part2}
\end{figure}

Figures~\ref{fig:old70-before-after-part1} and
\ref{fig:old70-before-after-part2} show all 70 original tasks for MACR (5M)
before and after 0.6M total new-task interactions (60k per new task).
We plot episodic returns; for Meta-World,
these differ from the success rates underlying the normalized retention scores.

\clearpage

\end{document}